%% file: 00-main.tex
\documentclass{IEEEtran}
\usepackage{amsfonts}
\input{01-config}

\title{In-Vivo Training for Deep Brain Stimulation \vspace*{-0.0cm}}
\author{Nicholas Carter$^1$ Arkaprava Gupta$^1$ Prateek Ganguli$^1$ Benedikt Dietrich$^2$ Vibhor Krishna$^3$ Samarjit Chakraborty$^1$ \\
$^1$Dept. of Computer Science, $^3$Dept. of Neurosurgery, UNC Chapel Hill, USA $^2$Hochschule M\"unchen, Germany \vspace*{-0.2cm}}

\begin{document}
\maketitle

\begin{abstract}
    Deep Brain Stimulation (DBS) is a highly effective treatment for Parkinson's Disease (PD). Recent research uses reinforcement learning (RL) for DBS, with RL agents modulating the stimulation frequency and amplitude. But, these models rely on biomarkers that are not measurable in patients and are only present in brain-on-chip (BoC) simulations. In this work, we present an RL-based DBS approach that adapts these stimulation parameters according to brain activity measurable \emph{in vivo}. Using a TD3 based RL agent trained on a model of the basal ganglia region of the brain, we see a greater suppression of biomarkers correlated with PD severity, compared to modern clinical DBS implementations. Our agent outperforms the standard clinical approaches in suppressing PD biomarkers while relying on information that can be measured in a real world environment, thereby opening up the possibility of training personalized RL agents specific to individual patient needs.
    \\[-0.5cm]  
\end{abstract}

\input{section/01-introduction}
\input{section/02-background}
\input{section/03-methodology}
\input{section/04-results}
\input{section/05-conclusion}

\vspace*{-.1cm}
\bibliographystyle{IEEEtran}
\bibliography{refs}

\end{document}

%% file: 01-config.tex
\usepackage{graphicx}
\usepackage{tikz}
\usepackage{enumitem}
\usepackage{wrapfig}
\usepackage{amsmath}
\usepackage[acronym]{glossaries}
\glsdisablehyper
\input{glossary}
\usetikzlibrary{arrows.meta, positioning, shapes.geometric}
\usepackage[justification=centering]{caption} 
\usepackage{booktabs}
\usepackage[capitalise]{cleveref}

%% file: glossary.tex
\newacronym{bgm}{BGM}{Basal Ganglia Model}

%% file: section/01-introduction.tex
\vspace*{-0.1cm}
\section{Introduction}

Deep Brain Stimulation (DBS) has been shown to be an effective treatment for Parkinson's Disease (PD). By providing electrical impulses to areas of the basal ganglia (BG) in the brain via a surgically inserted probe, the symptoms of PD can be mitigated. There are two main approaches to DBS: open-loop DBS (o-DBS) and closed-loop DBS (c-DBS). Open-loop DBS provides a constant frequency and stimulation to areas within the BG. However, it receives no feedback on the current state of the brain. This severely limits its adaptability, as the state of the brain is constantly changing based on new stimuli. Patients with o-DBS implants must regularly visit their neurologist, who modulates the stimulation parameters based on the observable health of the patient. Furthermore, the o-DBS system may deliver unneeded stimulations whenever the brain is in a resting state, wasting the battery life of the implant. Modulating the stimulation parameters based on the current needs of the brain is thus a much more efficient approach to DBS. This methodology encompasses c-DBS, where the probe sends adaptive stimulations based on biomarkers it measures, which are indicative of PD severity. \\[-.1cm]

\noindent 
\textbf{Related Works:}
Advances in deep learning and the inherent suitability of reinforcement learning for dynamic control problems have led to a number of studies that propose reinforcement learning (RL) implementations of c-DBS systems \cite{Cho, Lu, Gao, Gorzelic, Groiss}. Here, research has aimed to suppress the number of relay misses, denoted as the \emph{error index}~\cite{Cho}, in Parkinsonian thalamus neurons responding to input pulses from the sensorimotor cortex (SMC) in a computational model of the basal ganglia (BG) and the thalamus (TH). The work employed several RL agents in this task, namely a Twin Delayed Deep Deterministic Policy Gradient (TD3) agent, a Soft Actor-Critic (SAC) agent, a Proximal Policy Optimization (PPO) agent, and an Advantage Actor-Critic (A2C) agent, finding the TD3 agent to outperform at suppressing the error index while modulating the frequency and amplitude of the stimulation signal. Similar studies found that actor-critic based RL agents outperform standard o-DBS implementations in terms of power usage, while achieving similar PD severity biomarker reduction~\cite{Lu}. Combining AC algorithms with convolutional neural networks also prove to be effective c-DBS solutions~\cite{Gao}. Here, novel convolutional AC algorithm were used on a large array of various biomarkers, including the TH error index, and were found to outperform standard o-DBS implementations, while using less than a third of the energy. 
\\ [.2cm]
\textbf{Our Contributions:}
A majority of the RL implementations of c-DBS systems rely on the error index as a representative of the brain state. However, measuring the error index is not possible in a patient~\cite{Gorzelic}. This is because the SMC inputs to the TH will never be known, and thus calculating the error index via \emph{in vivo} measurements is impossible. Because of this, our work utilizes a biomarker more suitable for real world measurement. We choose the globus pallidus internus (GPi) synaptic conductance variable $S_{Gi}$ as our primary PD severity biomarker. The $S_{Gi}$ can be calculated via measurements of GPi voltage spike times obtained from local field potential values~\cite{Gorzelic}. Additionally, we measure brain health with the power of the beta band in GPi neurons, an indicator of PD intensity measurable in living brain tissues~\cite{Eisinger}. With these features, we use a TD3 RL approach similar to~\cite{Cho} and train the learning algorithm using a brain-on-chip (BoC) model of the BG and the thalamus, relying on metrics also obtainable in living brain tissues. These metrics are incorporated into a new feedback function, denoted as the \emph{reward} function in the RL parlance. It aids the algorithm in learning an appropriate stimulation strategy and parameters to alleviate PD severity. On completion, we obtain an RL based c-DBS system that reduced PD severity indicator and efficiently manages power. 

While RL has been studied for c-DBS in the past, the novelty of our approach lies in the use of new biomarkers and therefore a new reward function, and studying their effectiveness. In particular, we find that our TD3-DBS agent with these outperforms a standard o-DBS implementation~\cite{Cho}, reducing measured indicators of PD severity by an additional $7.35 \%$. Furthermore, our agent only consumes two-thirds of the energy utilized by o-DBS implementations~\cite{Cho}. Most importantly, our reliance on biomarkers measurable \emph{in vivo} is motivated by the possibility of continuing RL agent training after deployment in live patients. In this scenario, our BoC-trained learning agent would make small adjustments to its parameters to conform to the specific needs of any given patient, leading to a highly adaptive and personalized c-DBS system that requires no additional tuning from a neurologist. 


%% file: section/02-background.tex
\section{Background}\label{sec:background}



In this section, we briefly describe the BG-Thalamic (BGT) model we use for the implementation of our algorithm. The model consists of the thalamus (TH), the sub-thalamic nucleus (STN), the globus pallidus internus (GPi), and the globus pallidus externus (GPe). This model was first introduced by \cite{Rubin} and later improved by \cite{relative_contributions} to allow experimentation with DBS techniques for mitigation of PD. The BGT model is based on the ionic channel characteristics for the respective neuron types. For example, the following equation is used to model the dynamics of an single neuron of the thalamus:
$C_m \dot{v}_{TH} = -I_L - I_{Na} - I_K - I_T + I_{SMC} - I_{GPi \to TH},$
where $C_m$ is the membrane capacitance of , $\dot{v}_{TH}$ represents the time derivative of the membrane potential,  $I_L$, $I_{Na}$, $I_{K}$ and $I_T$ stand for leak, sodium, potassium and low-threshold calcium currents. $I_{SMC}$ is the current from the SMC and $I_{GPi \to TH}$ is the current from the GPi to the TH. The BGT model is able to mimic the functioning of both healthy and PD conditions of the brain. We use $10$ neurons for modeling each of the layers in the BGT model. It is understood that $10$ neurons can accurately simulate the functioning of the brain and the effect of DBS in the brain~\cite{relative_contributions}. DBS is generally preferred in the STN region in practice \cite{Kleiner}, and so we perform DBS in the same region in our implementation. \\[-0.15cm]

\noindent
\textbf{Biomarkers Used:} 
One of the main novelties of our method compared to previous work lies in the use of only those biomarkers that can be measured in-vivo in patients, opening up the possibility of in-vivo training of the RL agent, and therefore personalized treatment. We use the Power Spectral Density (PSD) of various neuronal signals to help measure PD intensity. We choose to measure the PSDs of the synaptic conductance of GPi neurons ($S_{Gi}$) and the membrane potential of GPi neurons ($V_{Gi}$). The GPi serves as the principal output nucleus of the BG, and its synaptic conductance plays a critical role in shaping BG output. This conductance state is dynamically modulated by dopamine, the key neurotransmitter deficient in PD. Alterations in dopamine tone influence excitatory and inhibitory synaptic inputs to GPi neurons, thereby affecting their firing patterns in patients with PD. These synaptic dynamics can be inferred from local field potential (LFP) recordings obtained through implanted deep brain stimulation electrodes in PD patients \cite{Gorzelic}. The PSD of the $13$ Hz $-$ $30$ Hz beta band measurable in GPi neurons is another biomarker correlated with PD intensity \cite{Eisinger}. \emph{In vivo}, this would be measured using LFP recordings similar to the $S_{Gi}$ readings. We calculate the PSD of a signal from $n$ neurons as follows: 
\\[-.2cm]
$$
\text{PSD}\left(x\right) = \frac{1}{n} \sum_{i=1}^{n} \int_{f_{\textbf{low}}}^{f_{\textbf{high}}} \left|\hat{f}\left(x_i\right)\right|\:df,
$$ 
\\[-.2cm]
where $\hat{f}(x)$ denotes the Fourier transform function and $x_i$ indicates the neuronal signal reading of the $i^\text{th}$ neuron from the GPi. When calculating the PSD of the $S_{Gi}$, we use $f_{\text{low}}= 1$ Hz and $f_{\text{high}}= 20$ Hz, whereas for the $V_{Gi}$, we use  $f_{\text{low}}= 13$ Hz and $f_{\text{high}}= 30$ Hz. In the BGT model, we observe that frequencies within these ranges are more prevalent in brains afflicted by PD compared to brains designated as healthy.  

%% file: section/03-methodology.tex
\section{Methodology}

\begin{wrapfigure}{r}{0.18\textwidth}
  \vspace{-.2in}
  \begin{center} 
    \hspace*{-0.1in}\includegraphics[width=0.22\textwidth]{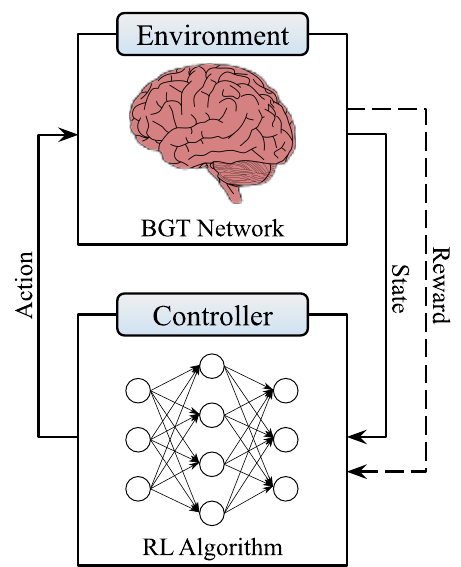}
  \end{center}
  \vspace{-.15in}
  \caption{\footnotesize The RL algorithm's interaction with the BGT model. The reward term is only used during training (dotted line).}
  \vspace{-.15in}
  \label{fig:brain}
\end{wrapfigure}

The formulation of our problem is outlined in Fig.~\ref{fig:brain}. We train an RL algorithm on the BGT environment to effectively reduce the biomarkers indicative of PD symptom severity. At specific time-steps, our RL agent monitors the \emph{state} of the brain, which is encapsulated in a vector of fixed size. Based on this state, the agent decides on an action it thinks best alleviates the PD severity biomarkers. Then, at the next time interval, it receives a reward based on how well its previous action suppressed these biomarkers. Using this feedback, it adjusts its internal decision making process, formulated as a collection neural networks. It then reads the state of the brain and selects another action, continuing this sequence of steps until it converges on a policy that is deemed optimal in mitigating these biomarkers. Following usual convention, we formulate our problem as a Markov Decision Process with the tuple $M = (\mathcal{S}, \mathcal{A}, \mathcal{R}, \mathcal{P})$. Here, $\mathcal{S}$ is a set representing the all possible states of the BGT environment, $\mathcal{A}$ is the set of actions the agent is allowed to make, $\mathcal{R}: \mathcal{S} \times \mathcal{A} \times \mathcal{S} \xrightarrow{} \mathbb{R}$ is the environment's innate reward function, and $P: \mathcal{S} \times \mathcal{A} \times \mathcal{S} \xrightarrow{} [0, 1]$ is a transition probability function. We now define how the state space, action space, and reward are defined in the BGT environment.

\subsubsection{State Space Representation}
We define the state space as a 6-element vector comprised the following:
\begin{itemize}[leftmargin=*]

    \item \textbf{The standard deviation of the $S_{Gi}$ signal} over the duration of the last timestep.
    
    \item \textbf{The Hjorth Parameters}, commonly used in the diagnoses of Alzheimer's disease and EEG based seizure detection \cite{Safi,Kaushik}. The Hjorth Parameters refer to the activity $A$, mobility $M$, and complexity $C$ of a given signal, and are defined by the following formulas: 
    \\[-0.2cm]
    \begin{equation*}
        A = \frac{1}{N} \sum_{i=1}^N (x_i - \bar{x})^2, \: \: M = \sqrt{\frac{A(\dot{x})}{A}}, \: \: C = \frac{M(\dot{x})}{M}.
    \end{equation*}
    \\[-0.2cm]
    Here, $N$ is the number of samples, $x$ is the $S_{Gi}$ signal, $\bar{x}$ is the mean of the signal, $\dot{x}$ is the time derivative of the signal.

    \item \textbf{The PSD of the membrane potential of the GPi neurons}, which encompasses brainwave activity in the $13-30 \text{ Hz}$ region, as mentioned in \cref{sec:background}. 

    \item \textbf{The sample entropy}, which measures the self-similarity of neuron readings in the STN. Parkinsonian brains exhibit higher levels of similarity \cite{Richman}, \cite{Fleming}, thus the sample entropy is inversely correlated to the PD severity. It is calculated as $S(m, r, N) = \text{ln}[C(m, r)]-\text{ln}[C(m+1, r)]$,
    where $m$ is the embedding dimension (default $m=2$), $r$ is the tolerance value for acceptance of the signal $x$ and is calculated as $0.2 \cdot \sigma_x$, where $\sigma_x$ is the standard deviation of $x$, and $N$ is the number of data samples. $C(a, b)$ denotes all embedded vectors of length $a$ with a Chebyshev distance lower than $b$. 
    \end{itemize}

    \noindent
    All state features are normalized between $[0, 1]$ using min-max normalization, preventing features of larger magnitude from dominating the learning process.   
    
    \subsubsection{Action Space Representation}
    We allow the RL agent to modulate both the frequency and amplitude of the DBS (current). This allows for finer grain control over the stimulations, leading to a much more adaptable DBS implementation than that of an open-loop system. The frequency and amplitude are continuous values within $[0, 185]$ Hz and $[0, 5000] \ \mu A/cm^2$, respectively. Each pulse has a width of 300 $\mu\text{s}$. Because RL models utilize Gaussian distributions centered at $0$ with a standard deviation of $-1$, when choosing continuous actions, we normalize the frequency and amplitude ranges to $[-1, 1]$. 
    Once the action is chosen by the RL agent, it is de-normalized and sent to the BGT environment. The DBS stimulation waveform is a symmetric, bi-phasic pulse with both anodic and cathodic stimulation. The anodic stimulation occurs first for $150$ $\mu\text{s}$ and is immediately followed by a $150$ $\mu\text{s}$ cathodic stimulation. A charge balanced bi-phasic current avoids unwanted faradic reactions that occur in the tissues surrounding the probe \cite{Piallat}.

    \subsubsection{Reward Representation}
    The agent interacts with the environment in discrete timesteps, receiving a reward based on the action it selected in the previous timestep and how the environment evolved as a result. The goal of the RL agent is to maximize its reward as it interacts with the environment. In the BGT environment, the agent should receive the reward based on how well it suppressed the 1 Hz to 20 Hz PSD of the $S_{Gi}$ signal while also using a minimal amount of power. Therefore, we designed our reward function around both factors. 
    As mentioned in \cref{sec:background}, the severity of PD based on the $S_{Gi}$ signal can be inferred by observing the $1$ Hz $-$ $20$ Hz range. We designate this factor of our reward function $r_1$. Power usage, denoted $r_2$, is calculated as $r_2 = \theta \cdot \frac{a_0+1}{2} + (1-\theta) \cdot \frac{a_1+1}{2}$, where $a_0$ and $a_1$ are the normalized frequency and amplitude chosen by the agent in the previous timestep, respectively. The hyperparameter $\theta \in [0, 1]$ is used to penalize either a high frequency or high amplitude. After some trial and error, we set $\theta$ to $0.85$ to dissuade the agent from selecting higher frequencies, as repeated high frequency stimulation can lead to hypophonia, oculomotor dysfunction, and nausea \cite{Groiss}. Prior work has not considered such issues in the design of the reward function, and hence this is another innovation in our work. 
    
    The overall reward is calculated via a weighted sum of the two reward factors as $r = \epsilon \cdot -r_1 + (1-\epsilon) \cdot -r2$. The $\epsilon$ term allows for prioritization of one factor over the other. In our case, we prioritized suppression of the $S_{Gi}$ PSD over power usage, and thus set $\epsilon = 0.68$. Both the $r_1$ and $r_2$ terms were multiplied by $-1$, as more positive values of both terms is unwanted; a higher $r_1$ indicates more severe PD and a higher $r_2$ indicates more power usage.

    \subsubsection{RL Model Selection}
    As the agent moves through the environment, it must learn the optimal policy to maximize the cumulative reward it receives at the end of each episode. The process by which the agent learns such a policy is determined by the learning methodology. For this experiment, we choose a Twin Delayed Deep Deterministic Policy Gradient (TD3) based RL agent based RL agent \cite{Fujimoto},  as \cite{Cho} found it to be the best performing RL algorithm on the BGT environment. But the success of this in our setting was not clear because of the new reward function we used. 
    The TD3 model is an actor-critic (AC) based learning method, where a neural network known as the \emph{actor} selects what it believes to be the optimal action given some environment state. This action is then evaluated by a neural network known as the \emph{critic}, which determines the future rewards to be gained by taking such an action in the given state. The actor and critic networks use each other to generate better estimations of both the optimal action and the expected rewards. Overestimating this future cumulative reward is a common issue in many AC based learning models. TD3 mitigates this issue by employing two critic networks and using the minimum of their estimated future rewards. 

    \subsubsection{Experimental Details}
    The RL agent interacts with the environment in discrete timesteps. It observes the state of the environment and selects an action it thinks is appropriate, carrying out that action until the next timestep. A shorter step length allows for more adaptability of the RL agent, but comes with the tradeoff of encompassing less information about the BGT state over the duration of the timestep. With this tradeoff in mind, we defined the length of one timestep to be $100 \text{ ms}$. Additionally, this ensures there will be at least one SMC pulse in each timestep. After a certain number of timesteps, the episode is expected to terminate. We capped our episode length at $1,000 \text{ ms}$, allowing $10$ timesteps per episode. Keeping the episode length relatively short allows our agent to experience a greater amount of episodes during training, leading to a more robust agent. Furthermore, we trained the agent for a maximum of $5,000$ timesteps, or terminated the training early when the average cumulative reward gained per episode converged. 
    \\[-0.7cm]

%% file: section/04-results.tex
\section{Experimental Results}


After completing the training process, we evaluated the performance of our TD3 agent in the BGT environment. We measured the average PSDs of the $S_{Gi}$ per timestep, that of the GPi neurons every 1 second, along with the average power usage.
Power usage was determined using the Root Mean Square (RMS) over the stimulation signal, calculated as $I_{RMS} = [\frac{1}{T}\int_0^TI_{DBS}^2(t)\ dt]^{\frac{1}{2}}$, with $I_{DBS}(t)$ being the amplitude of stimulation at time $t$. Power usage is thus measured in $\mu A \text{ cm}^{-2}\text{Hz}$. Using these metrics, we compared our TD3-DBS agent to a typical o-DBS system, which we assume to have a fixed frequency of 130 Hz and a fixed amplitude of 2500 $\mu A/\text{cm}^2$, which are parameters used in a very recent study~\cite{Cho}. Our comparison with the results from this o-DBS study is outlined in \cref{tab:comparison}. c-DBS experiments were reported in~\cite{Cho}, albeit using the \textit{error index} that cannot be computed using measurable biomarkers. Further, the formula for computing this error index was also not provided in~\cite{Cho}. We attempted to reproduce the c-DBS setup in~\cite{Cho} using the error index formula in~\cite{Lu}, but the resulting setup gave results that deviated significantly from the ones reported in~\cite{Cho}. Hence, we restricted our comparisons to o-DBS, which is not only common clinical practice but is also a prevalent point of comparison in multiple studies~\cite{Cho,Lu}. 

Our reward function was specifically designed to suppress the PSD of $S_{Gi}$, and with less importance minimize the energy consumption of the implant. We observe (\cref{tab:comparison}) that our resulting RL agent, denoted TD3-DBS, outperforms the o-DBS implementation in suppressing the PSDs of both, the $S_{Gi}$ and the $V_{Gi}$ of the GPi neurons. We see that our model achieves an average PSD $S_{Gi}$ value that is $7.35 \%$ less than that of the o-DBS average PSD $S_{Gi}$ value. Our model also achieves a $6.93\%$ reduction of the PSD $V_{Gi}$ compared to that of the o-DBS system. Even though the PSD $V_{Gi}$ term was not considered in the environment's reward function during training, it is still successfully suppressed during evaluation of the agent. We notice that while the TD3 agent adaptively chooses new stimulation parameters at each timestep, it gravitates towards frequencies similar to that of the o-DBS implementation. The average frequency and amplitude of stimulations provided by the TD3 agent were $135 \text{ Hz}$ and $1690$ $\mu A/\text{cm}^2$, respectively. While the average frequency was similar to the $130$ Hz used by o-DBS, the amplitude of stimulations in TD3-DBS was much lower. Therefore, the power usage of the TD3-DBS agent is  noticeably less than that of the o-DBS system. We note a power usage reduction from o-DBS to TD3-DBS of $31$$\%$. Despite a lower priority in the environment reward function compared to the reduction of $S_{Gi}$ PSD, the power usage by the agent was still managed efficiently. From these results, we conclude that our RL agent surpasses the performance of a typical o-DBS implementation, while only relying on biomarkers available in real world scenarios. We will make our code publicly available once the paper is accepted.  
\\[-0.7cm]

%% file: section/05-conclusion.tex
\section{Concluding Remarks \& Future Work}


\begin{table}[t]
  \centering
  \renewcommand{\arraystretch}{1.1}
\begin{tabular}{@{}c|cccc@{}}
\hline
\textbf{} & \textbf{$S_{Gi} \text{ PSD }$} &\textbf{$V_{Gi}$ PSD} & \textbf{Power Usage}\\
\textbf{} & \textbf{$(\mu V^2 Hz^{-1})$} &\textbf{$(\mu V^2$$Hz^{-1})$} & \textbf{$(\mu A $ $cm^{-2} Hz)$}\\ 
 \hline
 Healthy & $2200$ & $348000$ &  $-$\\
 PD & $3140$ & $896000$  & $-$\\ 
 \hline
 o-DBS & $1360$ & $361000$  & $494$\\ 
 \textbf{Our method} \\ (TD3-DBS) & $1260$ & $336000$ & $341$\\ 
 \hline
\end{tabular}
 \caption{Performance of our proposed RL agent. }    
  \label{tab:comparison}
\vspace{-0.7cm}
\end{table}

Current clinical DBS approaches employ fixed pattern stimulations to mitigate PD symptoms. Currently, patients with these implants must regularly visit a neurologist to have the stimulation parameters updated to best fit their needs. As the brain is constantly changing and exists in varying states, \textit{e.g.}, resting or awake, fixed stimulations lead to wasted battery life and patient discomfort. Thus, we proposed an RL based approach that adapts to patient needs, by varying stimulation parameters based on measurable biomarkers. Our TD3 RL agent modulates the frequency and amplitude of DBS stimulations every tenth of a second, leading to a high degree of adaptability between patients than an open-loop solution. We both, train and evaluate, our approach on a BoC simulation of the BG and TH regions, for use with RL algorithms. The simulation provides the RL agent with biological features also accessible in practical, real world scenarios. In doing so, we have prepared an RL agent more suitable for use in a DBS implementation, where these biomarkers will be recorded via brainwave activity from live tissue. In addition to rewarding the learning agent for suppressing PD indicative biomarkers, minimizing power usage also begets positive feedback. 
During evaluation of our TD3 based RL agent, we find it outperforms o-DBS in terms of minimizing PD indicative biomarkers in the BoC simulation, and does so utilizing considerably less energy. From this, we conclude that our reward function is effective at providing pertinent feedback on the health of the BoC and the overall energy usage. We also conclude that the biomarkers in the state representation of the BGT environment are effective indicators of brain health. The results presented in this work progress the knowledge of effective implementation of RL algorithms into realistic DBS scenarios, and away from metrics only obtainable in BoC models. We hope to further study the effects of more fine-grained time steps (from $100$ to $20$~ms) and changing the area of stimulation, possibly in the GPi instead of the STN. Additionally, we hope to make our RL agent usable outside of simulation, deploying it into a low latency real time system such as an FPGA. Furthermore, real time deployment will require safety mechanisms to prevent negative health effects caused by high frequency stimulation. We hope to maximize the adaptability and safety of our approach, contributing to the modernization of PD treatment and the advancement of personalized medicine. 
\\[-0.7cm]